\documentclass[12pt]{article}
\usepackage[letterpaper,margin=1in]{geometry}
\usepackage[T1]{fontenc}
\usepackage[utf8]{inputenc}
\usepackage{lmodern}
\usepackage{microtype}
\usepackage{graphicx}
\usepackage{booktabs}
\usepackage{array}
\usepackage{tabularx}
\usepackage{longtable}
\usepackage{xcolor}
\usepackage{amsmath,amssymb}
\usepackage[most]{tcolorbox}
\usepackage{enumitem}
\usepackage{float}
\usepackage{caption}
\usepackage{natbib}
\usepackage[colorlinks=true,linkcolor=blue,citecolor=blue,urlcolor=blue]{hyperref}

\setlist[itemize]{leftmargin=1.5em}
\setlist[enumerate]{leftmargin=1.5em}

\newcommand{\ATD}{\textit{Agentic Technical Debt}}
\newcommand{\ST}{\textit{Stochastic Tax}}
\newcommand{\Workflow}{\mathcal{W}}
\newcommand{\clip}{\Pi_{[0,1]}}
\newcommand{\E}{\mathbb{E}}

\title{Modeling Agentic Technical Debt and Stochastic Tax:\\A Standalone Framework for Measurement, Simulation, and Dashboarding}
\author{\begin{tabular}{c}
Muhammad Zia Hydari\\
School of Business, University of Pittsburgh\\
Pittsburgh, Pennsylvania, USA\\
\texttt{hydari@alum.mit.edu}\\[0.7em]
Raja Iqbal\\
Ejento.ai\\
Seattle, Washington, USA\\
\texttt{raja@ejento.ai}\\[0.7em]
Narayan Ramasubbu\\
School of Business, University of Pittsburgh\\
Pittsburgh, Pennsylvania, USA\\
\texttt{narayanr@pitt.edu}
\end{tabular}}
\date{}

\begin{document}
\maketitle

\begin{abstract}
Agentic AI systems combine probabilistic reasoning with delegated action through tools, context, memory, orchestration, and external workflow integration. This note develops a formal and managerially usable model that distinguishes \ATD{} from \ST{}. \ATD{} is a stock of accumulated design and governance liability. \ST{} is a recurring flow of operating burden that arises when stochastic agents are used in business workflows. The two constructs are related, but they are not the same: debt can amplify the tax, while the tax can remain positive even when debt is minimized. The note starts from a compact dashboard expression, expands it into a fuller structural model, defines all variables and parameters, shows how each cost category can be estimated from operational data, and illustrates the framework with an accounts-payable simulation and companion spreadsheet.
\end{abstract}

\section{Purpose and Core Intuition}

The purpose of this note is to make two constructs precise enough to support measurement, simulation, and dashboarding. Both constructs concern the management of agentic AI systems, but they answer different questions.

\begin{tcolorbox}[colback=blue!3,colframe=blue!50!black,title={Core distinction}]
\ATD{} is the accumulated liability created by expedient agent design and governance choices. \ST{} is the recurring cost of keeping stochastic agent behavior within acceptable operating bounds, and its rate can move with debt, usage, surface area, autonomy, and model variability.
\end{tcolorbox}

The distinction is most naturally expressed as a stock-flow distinction. \ATD{} is a stock because it accumulates over time through choices about prompts, tools, context, memory, routing, observability, and platform coupling. It can be paid down through refactoring, standardization, versioning, documentation, and governance. \ST{} is a flow because it is paid repeatedly as the organization operates an agentic workflow. It appears in evaluation, monitoring, retries, escalation, latency, token and context processing, guardrails, security checks, and revalidation.

This distinction also clarifies why the term \ST{} is useful. Interest on technical debt disappears when the debt is paid off. \ST{} need not disappear when \ATD{} is minimized. Even a carefully governed agentic workflow can impose a positive operating burden because probabilistic systems can vary across runs, act through tools, depend on context, and encounter new edge cases as adoption expands. Debt can raise this burden, but it does not create the burden from nothing.

\section{Conceptual Background}

The technical-debt metaphor originated in software engineering to describe how shortcuts can accelerate near-term delivery while increasing future maintenance cost \citep{cunningham1992wycash}. Later empirical work showed that technical debt can be measured through concrete liabilities and linked to remediation cost, operational disruption, and system longevity \citep{ramasubbu2015managing}. Machine-learning systems expanded the debt metaphor beyond code to include data dependencies, hidden feedback loops, pipeline entanglement, and undeclared consumers \citep{sculley2015hidden}. Foundation models and agentic systems extend the problem again because probabilistic outputs are no longer confined to prediction. They can guide tool calls, retrieve or write memory, trigger external actions, and revise plans across multi-step workflows \citep{bommasani2021opportunities,yao2022react}.

Traditional software systems are often deterministic: a defect may be expensive to fix, but it is usually reproducible. Predictive ML systems introduce probabilistic outputs, but the model is often embedded as a single inference component in a larger deterministic workflow. Agentic AI systems operate as control loops. They plan, call tools, condition on retrieved context, write or read memory, and act in external systems. This creates a distinctive managerial problem: the organization must govern probabilistic action, not only code quality or model accuracy.

\section{Definitions and Boundary Conditions}

Let $w \in \Workflow$ denote a workflow, such as accounts payable, customer support routing, or sales operations. Let $t$ denote a time period, such as a week, month, or release cycle.

Many of the same architectural surfaces appear in both definitions. Context handling, orchestration, memory, and tool routing can accumulate debt when teams make expedient choices; the same surfaces can also generate \ST{} because probabilistic execution varies across runs even when the design is disciplined.

\subsection{Agentic Technical Debt}

\textbf{Definition.} \ATD{} is the accumulated liability created when an agentic system is assembled, modified, or governed through expedient choices that make future change, validation, explanation, or control more costly than necessary.

The unit of liability is not only code or data. In agentic systems, debt also resides in instructions, context, state, tool contracts, action authority, and the governance routines that make future changes safe.

This liability can appear in some of the following places, including:

\begin{itemize}
    \item \textbf{Context and prompt debt:} long, inconsistent, or poorly versioned prompts; unstructured context; unclear retrieval rules; conflicting instructions.
    \item \textbf{Tool and schema debt:} unversioned tools, fragile tool-call schemas, undocumented connectors, inconsistent error handling, weak contracts between agents and tools.
    \item \textbf{Memory and state debt:} unclear write permissions, stale memory, weak state management, unclear retention policies, inconsistent memory retrieval.
    \item \textbf{Orchestration and routing debt:} brittle sequential chains, ad hoc routing logic, poorly specified escalation paths, untested workflow branches.
    \item \textbf{Governance and observability debt:} missing traces, weak golden sets, incomplete tests, insufficient monitoring, unclear ownership of prompt and tool changes.
    \item \textbf{Platform-coupling debt:} tight dependence on a particular model, vendor, safety policy, or connector interface without abstraction, versioning, or revalidation discipline.
\end{itemize}

\subsection{Stochastic Tax}

\textbf{Definition.} \ST{} is the recurring operating burden of keeping stochastic agentic behavior within acceptable bounds in a business workflow. The tax arises from the inherently probabilistic nature of agentic AI systems, not only from engineering shortcomings, and it can persist even if all technical debt were paid down. The relevant managerial question is therefore not whether the tax can be made to vanish, but which portion is an unavoidable cost of operating stochastic action and which portion is amplified by remediable debt.

This tax is distinct from the general operating burden of enterprise systems. ERP systems impose review delays, approval queues, and override workflows, but those costs are largely deterministic and predictable. \ST{} is specifically the variable burden introduced by non-determinism, namely the cost that exists because the same semantic input need not produce the same execution path.

\ST{} typically appears in the following recurring cost categories:

\begin{itemize}
    \item \textbf{Evaluation and test maintenance:} building and refreshing eval suites, running model or judge calls, and reviewing outputs as the workflow changes.
    \item \textbf{Monitoring and observability:} collecting traces, storing logs, maintaining dashboards, reviewing alerts, and investigating anomalous runs.
    \item \textbf{Retry and repair:} paying for additional model calls, tool calls, self-repair loops, and manual repair when the first execution path fails or becomes ambiguous.
    \item \textbf{Escalation and exception handling:} routing uncertain or risky cases to humans and absorbing the labor cost of review, approval, or correction.
    \item \textbf{Revalidation after change:} retesting the workflow after model, prompt, tool, context, policy, or vendor changes.
    \item \textbf{Latency and delay:} absorbing the cost of slower or more variable execution, especially when multi-step workflows create tail latency.
    \item \textbf{Token, compute, and context processing:} paying for longer prompts, larger retrieved context, repeated calls, model usage, compute, and tool-execution charges.
    \item \textbf{Security and guardrail maintenance:} updating input and output controls, reviewing security flags, maintaining guardrails, and testing new attack patterns.
\end{itemize}

\begin{tcolorbox}[colback=gray!4,colframe=gray!55!black,title={Illustrative boundary example},boxrule=0.4pt]
Suppose three users ask a general-purpose agentic assistant the same semantic question and need the same semantic answer. The model is strong and the surrounding scaffolding is mature. All three users may receive semantically similar outputs, but the number of intermediate reasoning steps, tool calls, retries, and tokens consumed can differ across the three runs. That cost variability is \ST{}: it arises from probabilistic execution, not from an error made by any of the users. By contrast, if one user has built a project with conflicting system prompts, undocumented connectors, and unclear context rules, that user is also carrying \ATD{}. The debt can be paid down by cleaning up prompts, versioning connectors, and documenting context policies. The tax can be reduced, but under current agentic architectures it cannot be assumed to disappear.

A smaller version of the same distinction appears when an agent searches the web before consulting a document that the user already uploaded. If the search fails because the article is paywalled and the agent then falls back to the uploaded document, the extra routing step is part of the stochastic operating burden. If the project repeatedly misroutes because its instructions and context hierarchy are poorly specified, that is also evidence of agentic debt.
\end{tcolorbox}

The term \ST{} carries two meanings, both of which the model preserves. First, it is a tax incurred \emph{because} the system is stochastic: probabilistic outputs, fallible tool calls, and a continuously expanding surface of edge cases create an operating burden that cannot be designed away under current agentic architectures. The model represents this reading through a baseline floor: even when agentic debt is set to zero, some evaluation, monitoring, guardrails, and exception handling remain. Second, the tax is \emph{itself} stochastic: its rate moves period to period as debt accumulates or is paid down, as adoption expands, as surface area grows, and as underlying event rates fluctuate. A team cannot quote next quarter's stochastic tax the way it quotes a lease payment. The formal model below makes both readings operational: the baseline floor captures the first, while the amplifiers and event-rate dependence capture the second.

Table~\ref{tab:contrast} summarizes the distinction.

\begin{table}[H]
\centering
\caption{Comparing \ATD{} and \ST{}}
\label{tab:contrast}
\small
\begin{tabularx}{\textwidth}{p{0.18\textwidth}p{0.38\textwidth}X}
\toprule
\textbf{Dimension} & \textbf{\ATD{}} & \textbf{\ST{}} \\
\midrule
Basic idea & Accumulated liability from expedient design and governance choices & Recurring operating burden of stochastic agent behavior \\
Economic metaphor & Debt principal & Tax or operating charge paid with use \\
Time pattern & Builds up over time & Paid per period, run, release, workflow, or transaction \\
Where it appears & Prompts, context, tools, schemas, memory, routing, orchestration, observability, platform coupling & Evaluation, monitoring, retries, escalation, revalidation, latency, token/context cost, security and guardrail work \\
Main question & What choices are making future change harder? & What recurring burden do we pay to operate behavior safely? \\
Managerial response & Refactor, standardize, version, document, redesign, govern & Measure, budget, set thresholds, automate checks, triage high-tax workflows \\
Can it be eliminated? & It can be reduced, sometimes materially & It can be reduced, but need not vanish under current agentic architectures \\
Relationship & Can amplify \ST{} & May reveal symptoms of \ATD{}, but is not reducible to \ATD{} \\
\bottomrule
\end{tabularx}
\end{table}

\section{Modeling \ATD{} as a Stock}

A useful model of \ATD{} must do two things. First, it must identify the components of debt that managers can inspect and remediate. Second, it must represent debt as persistent. A brittle prompt, undocumented connector, or unclear memory rule does not vanish merely because the next transaction succeeds.

\subsection{Debt Components}

For workflow $w$ in period $t$, define the debt-component vector
\begin{equation}
\mathbf{d}_{w,t}=\left(d^{ctx}_{w,t}, d^{tool}_{w,t}, d^{mem}_{w,t}, d^{orch}_{w,t}, d^{obs}_{w,t}, d^{plat}_{w,t}\right), \qquad d^i_{w,t}\in[0,1].
\label{eq:debt_vector}
\end{equation}

Each component $d^i_{w,t}$ is normalized between 0 and 1. A value near 0 means the component is well controlled. A value near 1 means the component creates severe future-change, validation, or control liability. Table~\ref{tab:atd_components} gives observable indicators for each component.

\begin{table}[H]
\centering
\caption{Operationalizing \ATD{} Components}
\label{tab:atd_components}
\small
\begin{tabularx}{\textwidth}{p{0.21\textwidth}p{0.31\textwidth}X}
\toprule
\textbf{Component} & \textbf{Observable indicators} & \textbf{Why it belongs in the debt model} \\
\midrule
Context and prompt debt & Prompt length growth, conflicting instructions, unversioned prompt edits, context-source sprawl & Prompt and context complexity make future changes hard to reason about and test. \\
Tool and schema debt & Unversioned tools, malformed tool calls, missing parameter checks, undocumented connectors & Agents act through tools, so weak contracts increase future validation and integration cost. \\
Memory and state debt & Stale memory, unclear write rules, missing retention policies, state inconsistencies & Persistent state lets past errors influence future behavior. \\
Orchestration and routing debt & Long serialized chains, fragile routing logic, excessive handoffs, unclear fallback paths & Multi-step workflows can compound errors and raise redesign cost. \\
Observability and governance debt & Missing traces, weak golden sets, unclear ownership, no release gates & Lack of observability makes future changes expensive and risky. \\
Platform-coupling debt & Tight dependence on a model, vendor API, safety policy, or connector without abstraction & External changes can force internal revalidation or redesign. \\
\bottomrule
\end{tabularx}
\end{table}

Equation~\ref{eq:debt_vector} is useful in practice because it gives the team a debt register. Each component can be scored from operational evidence, for example on a 0 to 5 scale and then normalized to $[0,1]$. The score need not be perfect to be useful. Its value comes from being repeated consistently over time and connected to remediation actions.

\subsection{Debt Accumulation Dynamics}

The next step is to model how a debt component changes over time. A component should decline when remediation is effective. It should rise from two sources: local change activity that the team accrues directly, and external shocks that the team absorbs from the platform. The local channel accrues debt only when there is something to change, the changes are made expediently, and governance is not strong enough to absorb them safely; we capture this with three multiplicative factors: change pressure $X$, shortcut intensity $Q$, and inverse governance $(1-G)$. The platform channel accrues debt only when the platform itself changes and the workflow is exposed to that change; we capture this with two multiplicative factors: platform volatility $V^{plat}$ and exposure $E$.

For component $i$, define
\begin{equation}
 d^i_{w,t+1}=\clip\left[(1-\phi_i R^i_{w,t})d^i_{w,t}+\alpha_i X^i_{w,t}Q^i_{w,t}(1-G^i_{w,t})+\zeta_i V^{plat}_{w,t}E^i_{w,t}\right].
\label{eq:debt_dynamics}
\end{equation}

The terms in Equation~\ref{eq:debt_dynamics} have the following meaning:

\begin{itemize}
    \item $d^i_{w,t}$ is the current debt score for component $i$ in workflow $w$.
    \item $R^i_{w,t}\in[0,1]$ is normalized remediation effort for component $i$, such as refactoring prompts, versioning tools, improving tests, or clarifying ownership.
    \item $\phi_i\ge 0$ converts remediation effort into debt reduction.
    \item $X^i_{w,t}\in[0,1]$ is change pressure on component $i$: the rate at which changes are demanded (prompt edits, new tools, new context sources, policy changes, or workflow changes) relative to the team's clean-change capacity during period $t$. High pressure forces edits to be made faster than they can be cleanly absorbed.
    \item $Q^i_{w,t}\in[0,1]$ is shortcut intensity, meaning the fraction of changes made through local patches rather than systematic redesign.
    \item $G^i_{w,t}\in[0,1]$ is governance maturity. Stronger versioning, tests, schemas, ownership, and observability reduce new debt creation.
    \item $V^{plat}_{w,t}\ge 0$ is platform volatility, such as changes in model behavior, vendor interfaces, safety policies, or connector APIs.
    \item $E^i_{w,t}\in[0,1]$ is exposure to platform volatility. A decoupled workflow has lower exposure than a workflow tightly coupled to one model or vendor.
    \item $\alpha_i$ and $\zeta_i$ translate local change pressure and platform volatility into debt growth.
    \item $\clip[\cdot]$ truncates the value to the unit interval.
\end{itemize}

Analytically, this equation captures persistence, accumulation, remediation, and platform-driven shocks. Practically, it tells a manager what to measure: change pressure, shortcut intensity, governance maturity, remediation effort, and exposure to external change. It also identifies levers. The team can reduce debt by increasing $R$, reducing $Q$, increasing $G$, or lowering platform exposure through abstraction and versioning.

The full equation is useful for diagnosing where debt comes from. For numerical exploration, however, managers often need a lighter recurrence that exposes only the amount of debt accumulated and the amount remediated during a period. The simulation and companion spreadsheet therefore use the following simplified per-period net-change rule:
\begin{equation}
D_{w,t+1}=\clip\left[D_{w,t}+\Delta^{acc}_{w,t}-\Delta^{rem}_{w,t}\right].
\label{eq:debt_dynamics_simple}
\end{equation}
Here $\Delta^{acc}_{w,t}$ is newly accumulated workflow-level debt during the period and $\Delta^{rem}_{w,t}$ is workflow-level remediation. This is the form used in the companion spreadsheet's \texttt{Time\_Series} sheet. It is a simplification of Equation~\ref{eq:debt_dynamics}, not a competing model.

\subsection{Aggregating Debt into a Workflow Index}

Managers need a workflow-level score in addition to component-level scores. A simple weighted average is often sufficient, but agentic components can interact. For example, prompt brittleness is more damaging when tool schemas are weak, and ungoverned memory is more damaging when routing is unclear. To capture this coupling, define
\begin{equation}
D_{w,t}=\clip\left[\sum_{i\in\mathcal{I}} \omega_i d^i_{w,t}+\sum_{i<j}\omega_{ij}d^i_{w,t}d^j_{w,t}\right].
\label{eq:debt_index}
\end{equation}

Here $D_{w,t}\in[0,1]$ is the normalized \ATD{} index, $\mathcal{I}$ is the set of debt components, $\omega_i\ge 0$ are component weights, and $\omega_{ij}\ge 0$ are coupling weights. A dashboard can report $100D_{w,t}$ as a 0 to 100 debt score. The interaction term is optional for a first implementation. A team with limited data can set all $\omega_{ij}=0$ and later add interactions once it observes that specific component pairs raise remediation cost or operating burden.

\subsection{Debt Principal in Dollars}

The index in Equation~\ref{eq:debt_index} is useful for tracking, but a financial view may require an estimated debt principal. If the team can estimate remediation costs, define
\begin{equation}
P^{ATD}_{w,t}=\sum_{i\in\mathcal{I}} \bar{c}^{rem}_i d^i_{w,t}+\sum_{i<j}\bar{c}^{coord}_{ij}d^i_{w,t}d^j_{w,t}+\bar{c}^{retest}_{w,t}.
\label{eq:debt_principal}
\end{equation}

$P^{ATD}_{w,t}$ is the estimated debt principal in dollars. $\bar{c}^{rem}_i$ is the estimated cost to fully remediate component $i$, $\bar{c}^{coord}_{ij}$ is the coordination premium created by coupled liabilities, and $\bar{c}^{retest}_{w,t}$ is the testing and revalidation cost required after remediation. This model is useful in practice because it translates an abstract score into a planning estimate: how much effort would it take to pay down the liability?

\section{Modeling \ST{} as a Flow}
\label{sec:st-flow}

A useful model of \ST{} must connect a dashboard-ready expression to observable operating drivers. This section starts with a compact average-cost expression, then expands each cost category into structural drivers that can be estimated, simulated, and implemented in a spreadsheet.

\subsection{From Dashboard Expression to Structural Expansion}

For workflow $w$ in period $t$, let $N_{w,t}$ denote completed transactions. Average stochastic tax per transaction can be expressed as recurring cost categories divided by completed transactions. In the expanded eight-category form used here,
\begin{equation}
\begin{split}
\overline{ST}_{w,t}=\frac{1}{N_{w,t}}\Big(&C^{eval}_{w,t}+C^{monitor}_{w,t}+C^{retry}_{w,t}+C^{escalate}_{w,t}+C^{revalidate}_{w,t}+C^{latency}_{w,t}\\
&+C^{token}_{w,t}+C^{security}_{w,t}\Big).
\end{split}
\label{eq:cacm_dashboard_expanded}
\end{equation}

In Equation~\ref{eq:cacm_dashboard_expanded}, $C^k_{w,t}$ is the total cost in category $k$ during period $t$. This expression is appropriate for a managerial dashboard because it reports a unit cost: the recurring burden per completed transaction. To populate the dashboard, however, each $C^k_{w,t}$ must be connected to observable drivers. The structural model below provides that expansion.

Table~\ref{tab:cost_categories} lists the eight cost categories used in the model.

\begin{table}[H]
\centering
\caption{Cost Categories for a \ST{} Dashboard}
\label{tab:cost_categories}
\small
\begin{tabularx}{\textwidth}{p{0.23\textwidth}p{0.32\textwidth}X}
\toprule
\textbf{Cost category $k$} & \textbf{Measurement approach} & \textbf{Typical data source} \\
\midrule
Evaluation and test maintenance & Model calls, judge calls, test infrastructure, expert review time & Evaluation logs, test records, reviewer hours \\
Monitoring and observability & Dashboard cost, trace storage, alert review, routine operations time & Observability tools, log storage, incident queues \\
Retry and repair & Extra model calls, extra tool calls, self-repair loops, manual repair work & Agent traces, tool-call logs, retry counters \\
Escalation and exception handling & Escalated cases times average handling time and labor rate & Ticket queues, approval logs, operations records \\
Revalidation after change & Retesting and approval after model, prompt, tool, policy, or context changes & Release records, model-version logs, validation plans \\
Latency and delay & Excess latency relative to target times delay cost, plus SLA or abandonment cost & Latency dashboards, SLA records, conversion or productivity estimates \\
Token, compute, and context processing & Expected tokens per transaction times unit cost, plus compute and tool-call charges & Model bills, token logs, context retrieval logs \\
Security and guardrail maintenance & Security flags, input/output guardrail reviews, adversarial test updates & Security logs, guardrail logs, red-team records \\
\bottomrule
\end{tabularx}
\end{table}

\subsection{Total and Average Stochastic Tax}

Let $TST_{w,t}$ denote total stochastic tax for workflow $w$ in period $t$. Let $\mathcal{K}$ denote the eight cost categories in Table~\ref{tab:cost_categories}. Then
\begin{equation}
TST_{w,t}=\sum_{k\in\mathcal{K}} C^k_{w,t}.
\label{eq:total_st}
\end{equation}

The per-transaction tax reported on a dashboard is
\begin{equation}
\overline{ST}_{w,t}=\frac{TST_{w,t}}{N_{w,t}}.
\label{eq:avg_st}
\end{equation}

Equations~\ref{eq:total_st} and \ref{eq:avg_st} separate total budget burden from unit economics. Total tax may rise with adoption even while per-transaction tax falls because fixed evaluation, monitoring, and governance costs are spread over more transactions. This distinction is important for managers because a scaled agent can look more expensive in total dollars while becoming more efficient per transaction.

\subsection{Scales for Operating Drivers}

Before specifying the structural form, the operating drivers need clear scales. Table~\ref{tab:driver_scales} defines the variables used in the operating-exposure amplifier.

\begin{table}[H]
\centering
\caption{Operating-Exposure Variables and Scales}
\label{tab:driver_scales}
\small
\begin{tabularx}{\textwidth}{p{0.12\textwidth}p{0.27\textwidth}X}
\toprule
\textbf{Symbol} & \textbf{Scale} & \textbf{Interpretation} \\
\midrule
$U_{w,t}$ & Positive count or index & Adoption or usage exposure, such as transactions, active users, teams, or distinct usage contexts. In the simulation, $U=N$. \\
$S_{w,t}$ & Positive count or index & Surface area, such as tools, connectors, context sources, permissions, APIs, or external systems. \\
$H_{w,t}$ & Positive count & Workflow horizon, measured as the number of dependent agentic steps or the effective depth of the action chain. \\
$A_{w,t}$ & $[0,1]$ & Autonomy and action criticality: 0 = advisory, 0.5 = approval-required, 1 = direct execution in external systems. \\
$\Theta_{w,t}$ & $[0,1]$ & Model or platform variability: 0 = stable behavior under fixed tests, 0.5 = moderate run-to-run or version drift, 1 = high variability or rapid provider-side change. \\
\bottomrule
\end{tabularx}
\end{table}

These scales make the coefficients interpretable. A manager can score $A$ and $\Theta$ consistently across workflows even when the exact engineering measures differ.

\subsection{Cost-Category Structural Form}

Each cost category has three drivers: a baseline floor, a debt amplifier, and an operating-exposure amplifier. The baseline floor is what the workflow would cost at zero \ATD{} and at reference operating conditions. The debt amplifier, $\Phi_k(D_{w,t})$, is a function of $D_{w,t}$ and captures the effect of accumulated \ATD{}. The operating-exposure amplifier, $\Psi_k(U_{w,t},S_{w,t},H_{w,t},A_{w,t},\Theta_{w,t})$, captures adoption, surface area, workflow horizon, autonomy, and inherent model variability.

For each category $k$, define
\begin{equation}
C^k_{w,t}=\underbrace{\left(F_k+V_kN_{w,t}\right)}_{\text{baseline floor}}\cdot \underbrace{\Phi_k(D_{w,t})}_{\text{debt amplifier}}\cdot \underbrace{\Psi_k(U_{w,t},S_{w,t},H_{w,t},A_{w,t},\Theta_{w,t})}_{\text{operating-exposure amplifier}}.
\label{eq:cost_factorization}
\end{equation}

The variables and parameters in Equation~\ref{eq:cost_factorization} are:

\begin{itemize}
    \item $F_k\ge 0$ is the fixed or semi-fixed cost of category $k$ in a period, such as maintaining an evaluation suite or observability pipeline.
    \item $V_k\ge 0$ is the baseline variable cost per completed transaction in category $k$.
    \item $N_{w,t}$ is completed transactions.
    \item $D_{w,t}$ is the normalized \ATD{} index from Equation~\ref{eq:debt_index}.
    \item $U_{w,t}$, $S_{w,t}$, $H_{w,t}$, $A_{w,t}$, and $\Theta_{w,t}$ are the operating-exposure variables defined in Table~\ref{tab:driver_scales}.
    \item $\Phi_k(\cdot)$ and $\Psi_k(\cdot)$ are normalized amplifiers defined below.
\end{itemize}

Analytically, Equation~\ref{eq:cost_factorization} makes the floor visible. The model cannot accidentally say that operating cost is zero when debt is zero. Practically, the equation gives managers a way to ask whether a rise in cost is coming from more usage, a larger surface area, longer workflows, greater autonomy, external variability, or accumulated debt.

\subsection{Debt Amplifier}

The simplest useful debt amplifier is linear:
\begin{equation}
\Phi_k(D_{w,t})=1+\beta_kD_{w,t}, \qquad \beta_k\ge 0, \qquad \Phi_k(0)=1.
\label{eq:phi}
\end{equation}

The coefficient $\beta_k$ is the debt sensitivity of category $k$. If $\beta_{retry}=2.5$, then moving from $D=0$ to $D=1$ multiplies retry-related cost by $3.5$, holding other drivers fixed. Larger values of $\beta_k$ are plausible for retry, revalidation, and evaluation because brittle prompts, weak tool schemas, and poor observability tend to appear first in those categories. Smaller values may be plausible for categories that are more infrastructure-driven, such as baseline monitoring.

The practical value of $\beta_k$ depends on calibration. A team can begin with expert estimates, but the debt-amplified decomposition should be treated as a scenario estimate until the team has observed enough periods, incidents, releases, or remediation interventions to update the values.

\subsection{Operating-Exposure Amplifier}

The operating-exposure amplifier captures the fact that \ST{} can rise even when \ATD{} is unchanged. More adoption means more users, prompts, edge cases, and exception opportunities. More surface area means more tools, context sources, permissions, and possible attack vectors. Longer horizons make cascading errors more likely. Higher autonomy increases the need for control. Greater model variability raises the need for evaluation and revalidation.

Let $U_0$, $S_0$, $H_0$, $A_0$, and $\Theta_0$ denote reference operating values. A coherent positive form is
\begin{align}
\Psi_k(U_{w,t},S_{w,t},H_{w,t},A_{w,t},\Theta_{w,t}) &= \exp\Bigg[\gamma^U_k\ln\left(\frac{U_{w,t}}{U_0}\right)+\gamma^S_k\ln\left(\frac{S_{w,t}}{S_0}\right) \nonumber \\
&\quad +\gamma^H_k\ln\left(\frac{H_{w,t}}{H_0}\right)+\gamma^A_k(A_{w,t}-A_0)+\gamma^\Theta_k(\Theta_{w,t}-\Theta_0)\Bigg].
\label{eq:psi}
\end{align}

Here $\gamma^U_k$, $\gamma^S_k$, and $\gamma^H_k$ are elasticities with respect to adoption, surface area, and horizon. For example, $\gamma^U_k=0.10$ means that a tenfold increase in $U$ multiplies category-$k$ cost by approximately $10^{0.10}$, holding other drivers fixed. $\gamma^A_k$ and $\gamma^\Theta_k$ are semi-elasticities for the bounded scores $A$ and $\Theta$. The exponential form guarantees that $\Psi_k>0$ without imposing an arbitrary floor. At the reference values, $\Psi_k=1$.

For dashboard implementation, teams can start with only $U$ and $S$ if data are limited, then add $H$, $A$, and $\Theta$ as the measurement program matures.

\subsection{Three Properties Built Into the Model}

The model encodes three properties that clarify why \ST{} is not simply interest on \ATD{}.

\begin{tcolorbox}[enhanced,colback=blue!4,colframe=blue!50!black,title={Three properties of the model},boxrule=0.4pt]
\textbf{P1. Nonzero baseline.} When $D_{w,t}=0$, $\Phi_k(0)=1$, but $TST_{w,t}$ remains positive if at least one category has $F_k+V_kN_{w,t}>0$. A zero-debt agentic workflow still carries stochastic tax. This is the formal statement of the first meaning of \ST{}: the tax exists because the system is stochastic, not because the workflow is badly engineered.

\medskip
\textbf{P2. Debt amplification.} If $\beta_k\ge 0$, then
\[
\frac{\partial C^k_{w,t}}{\partial D_{w,t}}=(F_k+V_kN_{w,t})\beta_k\Psi_k(\cdot)\ge 0.
\]
Accumulated debt raises category cost. Debt amplifies the tax; it does not create the tax from nothing.

\medskip
\textbf{P3. Independent operating-exposure channel.} If any operating-exposure coefficient is positive, then \ST{} can rise even when $D_{w,t}$ is unchanged. Adoption, surface area, workflow horizon, autonomy, and model variability can each raise the tax independently. P2 and P3 formalize the second meaning of \ST{}: the tax rate itself moves with debt, operating exposure, and the event rates that those drivers influence.
\end{tcolorbox}

A useful implication follows. If a team pays down \ATD{} but still observes a rising \ST{}, this is not necessarily a failure of debt remediation. It may mean the system is being used more widely, exposes more tools and data sources, or operates with higher autonomy. The dashboard should therefore decompose observed tax rather than treating every increase as evidence of debt.

\subsection{Baseline and Debt-Amplified Tax}

The conceptual dashboard decomposition is
\begin{equation}
\overline{ST}_{w,t}=\overline{ST}^{0}_{w,t}+\overline{ST}^{D}_{w,t},
\label{eq:decomp}
\end{equation}
where
\begin{equation}
\overline{ST}^{0}_{w,t}=\overline{ST}_{w,t}\mid D_{w,t}=0,
\label{eq:baseline}
\end{equation}
\begin{equation}
\overline{ST}^{D}_{w,t}=\overline{ST}_{w,t}-\overline{ST}^{0}_{w,t}.
\label{eq:debt_amplified}
\end{equation}

$\overline{ST}^{0}_{w,t}$ is the baseline stochastic tax, or the tax associated with operating a stochastic, tool-using workflow at the current level of adoption, surface area, horizon, autonomy, and variability. $\overline{ST}^{D}_{w,t}$ is the debt-amplified tax, or the additional burden associated with accumulated \ATD{}.

This decomposition is conceptually important, but it should not be overinterpreted before calibration. Computing $\overline{ST}^{0}_{w,t}$ requires values for $\beta_k$. Those values are usually not directly observed. A manager should therefore report the decomposition as a calibrated estimate, with sensitivity ranges, until the organization has learned $\beta_k$ from history, interventions, or expert judgment.

\subsection{Calibrating Debt Sensitivities}

A feasible way to calibrate $\beta_k$ is to compare periods or workflows with materially different debt scores while holding operating exposure approximately constant. Define the normalized observed category cost
\begin{equation}
Z^k_{w,t}=\frac{C^k_{w,t}}{(F_k+V_kN_{w,t})\Psi_k(U_{w,t},S_{w,t},H_{w,t},A_{w,t},\Theta_{w,t})}.
\label{eq:z_norm}
\end{equation}

Under Equation~\ref{eq:cost_factorization}, $Z^k_{w,t}=1+\beta_kD_{w,t}$. With two observations $t_1$ and $t_2$,
\begin{equation}
\widehat{\beta}_k\approx \frac{Z^k_{w,t_2}-Z^k_{w,t_1}}{D_{w,t_2}-D_{w,t_1}}, \qquad D_{w,t_2}\ne D_{w,t_1}.
\label{eq:beta_calibration}
\end{equation}

In practice, a team might estimate $\widehat{\beta}_k$ after a refactoring sprint that lowers tool/schema debt, after a prompt rationalization project, or across comparable workflows with different debt scores. If such observations are not available, $\beta_k$ should be treated as an expert prior and stress-tested in the spreadsheet. The decomposition in Equations~\ref{eq:decomp} to \ref{eq:debt_amplified} becomes more decision-ready as these estimates improve.

Other parameters can be estimated with the same pragmatic discipline. The fixed terms $F_k$ and baseline unit costs $V_k$ can come from accounting allocations, model bills, observability invoices, labor rates, and logs at a reference operating point. The exposure coefficients $\gamma$ can be updated from before-after observations when adoption, surface area, horizon, autonomy, or model variability changes materially while other drivers remain approximately stable. Unit costs such as $c^{tok}_t$ and $c^{lat}_{w,t}$ can come from provider pricing, internal compute charges, labor-cost models, SLA penalties, or business estimates of delay. Event rates such as $q^j_{w,t}$ can be read from traces, incident queues, security logs, and escalation records. This keeps the model feasible: teams can begin with approximate but transparent assumptions, then replace them with observed estimates over time.

\section{Estimating Cost Categories from Operational Data}

The structural model explains and forecasts category costs. A dashboard also needs measurement rules for estimating each $C^k_{w,t}$ from logs, invoices, tickets, and labor records. These measurement equations produce the same quantities as the structural model when calibrated to the same drivers; they differ in inputs. The structural form asks, ``what should the cost be, given debt and operating exposure?'' The measurement layer asks, ``what did the cost actually amount to this period?''

A general direct-measurement expression is
\begin{equation}
C^k_{w,t}=F^{obs,k}_{w,t}+\sum_{m\in\mathcal{M}_k} x^{k,m}_{w,t}p^{k,m}_{w,t}.
\label{eq:direct_measurement}
\end{equation}
Here $F^{obs,k}_{w,t}$ is an observed fixed or allocated cost for category $k$, $\mathcal{M}_k$ is the set of measured resource types in that category, $x^{k,m}_{w,t}$ is the observed quantity of resource $m$ used in category $k$, and $p^{k,m}_{w,t}$ is the unit price of that resource. Examples include model calls, reviewer hours, storage volume, security analyst time, tool calls, and escalated tickets.

For rate-based categories such as retry, escalation, and security flags, a convenient special case is
\begin{equation}
C^j_{w,t}=N_{w,t}q^j_{w,t}c^j_{w,t},
\label{eq:event_cost}
\end{equation}
where $j$ is an event type, $q^j_{w,t}$ is the event rate per transaction, and $c^j_{w,t}$ is the average cost per event. For example, escalation cost can be estimated as transactions times escalation rate times average handling cost.

Token and context-processing costs can often be measured directly from model logs:
\begin{equation}
C^{token}_{w,t}=N_{w,t}\frac{\E[Tok_{w,t}]}{1000}c^{tok}_{t}+C^{compute}_{w,t}.
\label{eq:token_cost}
\end{equation}
$\E[Tok_{w,t}]$ is expected tokens consumed per transaction, $c^{tok}_{t}$ is the blended cost per 1,000 tokens, and $C^{compute}_{w,t}$ captures additional compute, retrieval, or tool-execution charges not included in the token price. In the structural model, the baseline value of this expression at reference operating conditions is absorbed into $F_{token}+V_{token}N$.

Latency cost can be measured relative to a target:
\begin{equation}
C^{latency}_{w,t}=N_{w,t}c^{lat}_{w,t}\E\left[(L_{w,t}-L^*_w)_+\right].
\label{eq:latency_cost}
\end{equation}
Here $L_{w,t}$ is observed latency, $L^*_w$ is the target latency for workflow $w$, $(x)_+=\max(0,x)$, and $c^{lat}_{w,t}$ is the business cost of one unit of excess delay. As with token cost, the structural parameter $V_{latency}$ absorbs the expected reference-condition cost, while the observed measurement equation is used to populate and recalibrate the dashboard.

Table~\ref{tab:measurement_rules} links each category to a practical measurement rule.

\begin{table}[H]
\centering
\caption{Practical Measurement Rules for Cost Categories}
\label{tab:measurement_rules}
\small
\begin{tabularx}{\textwidth}{p{0.22\textwidth}p{0.38\textwidth}X}
\toprule
\textbf{Category} & \textbf{First-cut measurement} & \textbf{Common pitfall} \\
\midrule
Evaluation & Test runs, judge calls, and reviewer hours times unit costs & Forgetting maintenance of the test set itself. \\
Monitoring & Allocated observability cost plus alert review hours & Allocating shared monitoring cost inconsistently across workflows. \\
Retry and repair & Retry count times incremental model/tool cost plus repair labor & Counting retries but ignoring self-repair loops that still consume tokens. \\
Escalation & Escalation rate times average handling cost & Treating escalations as failures even when they are appropriate controls. \\
Revalidation & Retest hours and calls after model, prompt, tool, context, or policy changes & Charging all revalidation to the model rather than the workflow that required it. \\
Latency and delay & Expected excess latency times business delay cost & Using average latency while ignoring tail latency. \\
Token, compute, and context & Token logs, model bills, retrieval charges, and tool-execution cost & Ignoring context retrieval and repeated calls. \\
Security and guardrails & Flag review time, red-team updates, guardrail calls, and security operations effort & Treating guardrails as one-time implementation rather than recurring maintenance. \\
\bottomrule
\end{tabularx}
\end{table}

The measurement layer and structural layer should be used together. Observed category costs provide the dashboard numbers. The structural model explains why those numbers change and supports simulations of what could happen if debt, adoption, surface area, horizon, autonomy, or model variability changes.

\section{Numerical Illustration: Accounts-Payable Agent}

The simulation below illustrates the model with a hypothetical accounts-payable agentic workflow. The workflow extracts invoice fields, checks vendor identity, validates payment terms, applies policy rules, and schedules payment. It uses several tools and context sources, produces traces, and escalates exceptions to a human queue. The calibration is illustrative. It is designed to be plausible enough for managerial reasoning, not to estimate a universal industry benchmark.

\subsection{Cost-Category Calibration}

Table~\ref{tab:calib} reports the primary cost parameters. $V_k$ is baseline variable cost per transaction. $F_k$ is fixed monthly cost. $\beta_k$ is the debt amplifier. The $\gamma$ columns are operating-exposure sensitivities. All dollar values are monthly and illustrative.

\begin{table}[H]
\centering
\caption{Cost-Category Parameters for the Accounts-Payable Simulation}
\label{tab:calib}
\footnotesize
\begin{tabularx}{\textwidth}{p{0.22\textwidth}rrrrrrrX}
\toprule
\textbf{Category $k$} & $V_k$ & $F_k$ & $\beta_k$ & $\gamma^U_k$ & $\gamma^S_k$ & $\gamma^H_k$ & $\gamma^A_k$ & \textbf{Main intuition} \\
\midrule
Evaluation & 0.010 & 200 & 1.2 & 0.12 & 0.08 & 0.08 & 0.12 & Fragile systems need more test and reviewer effort. \\
Monitoring & 0.005 & 350 & 0.4 & 0.10 & 0.06 & 0.04 & 0.06 & Trace volume and alert triage rise with scale. \\
Retry and repair & 0.040 & 0 & 2.5 & 0.07 & 0.05 & 0.12 & 0.10 & Brittle prompts and schemas produce malformed calls. \\
Escalation & 0.120 & 80 & 1.0 & 0.08 & 0.06 & 0.10 & 0.18 & Exceptions need human review. \\
Revalidation & 0.002 & 400 & 1.8 & 0.10 & 0.12 & 0.12 & 0.10 & Model, tool, and policy changes need retesting. \\
Latency and delay & 0.030 & 50 & 0.8 & 0.05 & 0.05 & 0.18 & 0.15 & Long chains and retries create delay. \\
Token, compute, and context & 0.200 & 0 & 0.6 & 0.03 & 0.08 & 0.12 & 0.08 & Large context and repeated calls raise compute cost. \\
Security and guardrails & 0.008 & 150 & 0.3 & 0.15 & 0.18 & 0.05 & 0.20 & More tools and users expand attack surface. \\
\bottomrule
\end{tabularx}
\end{table}

The corresponding $\gamma^\Theta_k$ values are 0.15, 0.08, 0.12, 0.08, 0.20, 0.10, 0.08, and 0.12 for evaluation, monitoring, retry and repair, escalation, revalidation, latency and delay, token/context, and security/guardrails, respectively. The companion spreadsheet exposes all parameters as editable cells.

\subsection{Four-Scenario Comparative Statics}

The first simulation uses a $2\times2$ design: low versus high adoption crossed with low versus high debt. Holding other variables constant makes the interpretation clean. It shows what scale does when debt is low, what debt does when adoption is low, and what happens when scale and debt meet.

\begin{table}[H]
\centering
\caption{Scenario Inputs}
\label{tab:scenario_inputs}
\footnotesize
\begin{tabularx}{\textwidth}{lrrrrrrX}
\toprule
\textbf{Scenario} & $N$ & $D$ & $U$ & $S$ & $H$ & $A$ & \textbf{Description} \\
\midrule
S1: Low adoption, low debt & 1,000 & 0.10 & 1,000 & 5 & 4 & 0.40 & Carefully governed pilot. \\
S2: High adoption, low debt & 10,000 & 0.10 & 10,000 & 15 & 4 & 0.40 & Scaled with disciplined governance. \\
S3: Low adoption, high debt & 1,000 & 0.60 & 1,000 & 5 & 4 & 0.40 & Pilot with accumulated debt. \\
S4: High adoption, high debt & 10,000 & 0.60 & 10,000 & 15 & 4 & 0.40 & Scaled with accumulated debt. \\
\bottomrule
\end{tabularx}
\end{table}

All four scenarios set $\Theta=0.30$. This keeps the example focused on adoption, surface area, and debt.

\begin{table}[H]
\centering
\caption{Stochastic Tax Under the Four Scenarios}
\label{tab:scenario_results}
\footnotesize
\begin{tabularx}{\textwidth}{lrrrr}
\toprule
\textbf{Scenario} & \textbf{Total TST} & \textbf{$\overline{ST}$ per tx} & \textbf{Baseline per tx} & \textbf{Debt-amplified per tx} \\
\midrule
S1: Low adoption, low debt & \$1,810 & \$1.81 & \$1.65 & \$0.16 \\
S2: High adoption, low debt & \$7,520 & \$0.75 & \$0.69 & \$0.07 \\
S3: Low adoption, high debt & \$2,634 & \$2.63 & \$1.65 & \$0.99 \\
S4: High adoption, high debt & \$10,799 & \$1.08 & \$0.69 & \$0.39 \\
\bottomrule
\end{tabularx}
\end{table}

Three patterns are visible in Table~\ref{tab:scenario_results}. First, the tax is positive even at low debt. Second, adoption can reduce per-transaction tax by amortizing fixed evaluation, monitoring, and governance costs. In the low-debt case, scaling from S1 to S2 reduces per-transaction tax by about \$1.06. Third, debt steals some of these economies of scale. Scaling from S1 to S4 reduces per-transaction tax by only about \$0.73. The difference, roughly \$0.33 per transaction, is a useful managerial number: it is the per-transaction saving that accumulated debt prevents the team from capturing.

\begin{figure}[H]
\centering
\includegraphics[width=0.82\textwidth]{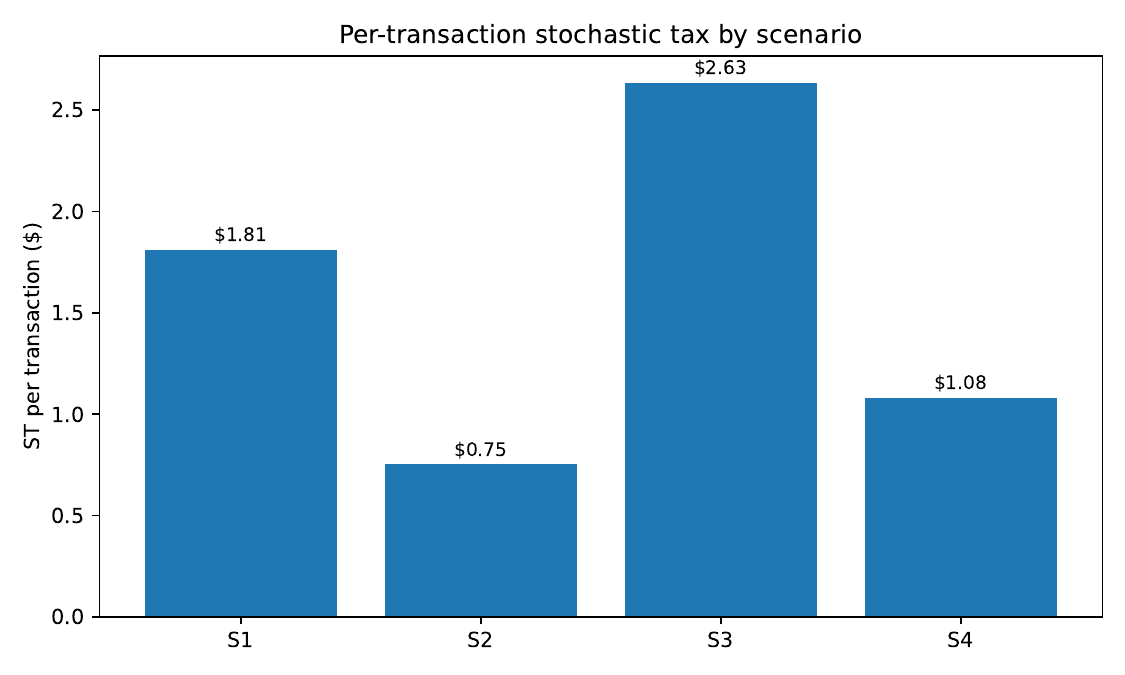}
\caption{Per-transaction \ST{} across the four scenarios.}
\label{fig:scenario_per_tx}
\end{figure}

\begin{figure}[H]
\centering
\includegraphics[width=0.82\textwidth]{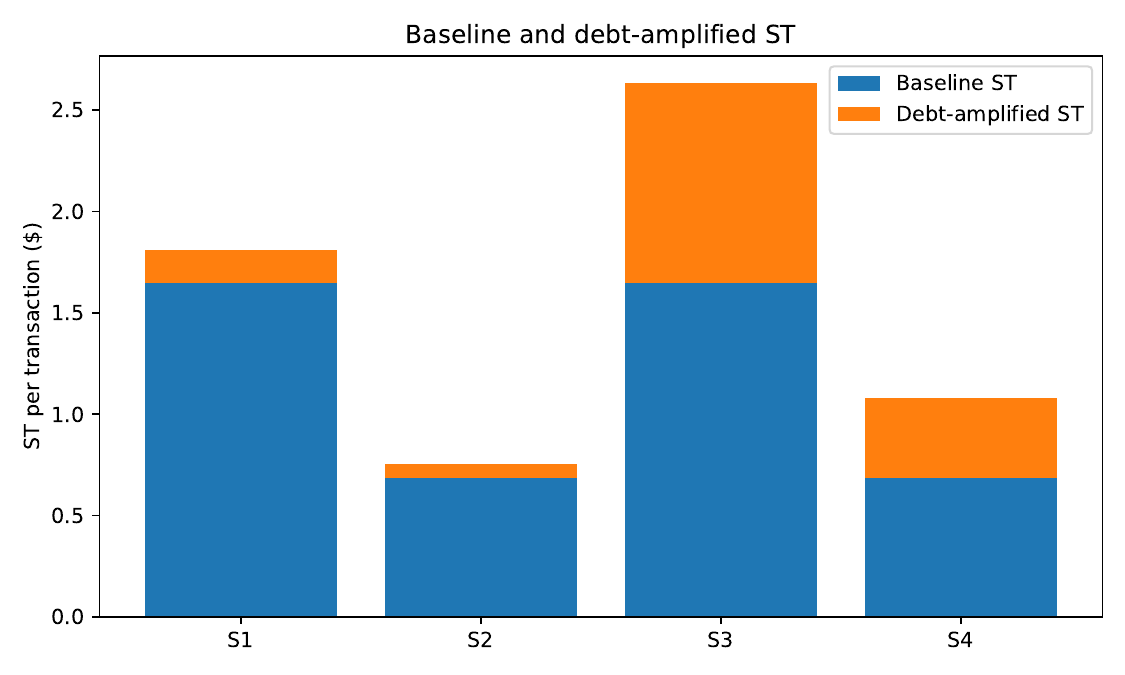}
\caption{Baseline and debt-amplified \ST{} by scenario. The baseline component remains positive when debt is set to zero.}
\label{fig:decomposition}
\end{figure}

\subsection{ATD Sensitivity Sweep}

The second simulation isolates the debt channel. It holds adoption, surface area, horizon, autonomy, and model variability fixed at the high-adoption operating point and varies $D$ from 0 to 1. This answers a practical question: if our debt score rises by 0.1 at the current scale, how much will per-transaction tax increase?

\begin{table}[H]
\centering
\caption{Sensitivity of \ST{} to the \ATD{} Index at High Adoption}
\label{tab:atd_sensitivity}
\small
\begin{tabularx}{\textwidth}{rrrr}
\toprule
$D$ & \textbf{Total TST} & \textbf{$\overline{ST}$ per tx} & \textbf{Change vs. $D=0$} \\
\midrule
0.00 & \$6,864 & \$0.69 & -- \\
0.25 & \$8,504 & \$0.85 & +24\% \\
0.50 & \$10,143 & \$1.01 & +48\% \\
0.75 & \$11,783 & \$1.18 & +72\% \\
1.00 & \$13,423 & \$1.34 & +96\% \\
\bottomrule
\end{tabularx}
\end{table}

At $D=0$, the per-transaction tax is still \$0.69. This is the nonzero floor. At this operating point, the relationship is approximately linear because the debt amplifier is linear. The local slope is managerially interpretable: it is the expected per-transaction increase in \ST{} for a unit increase in the debt index, holding other operating conditions fixed.

\begin{figure}[H]
\centering
\includegraphics[width=0.82\textwidth]{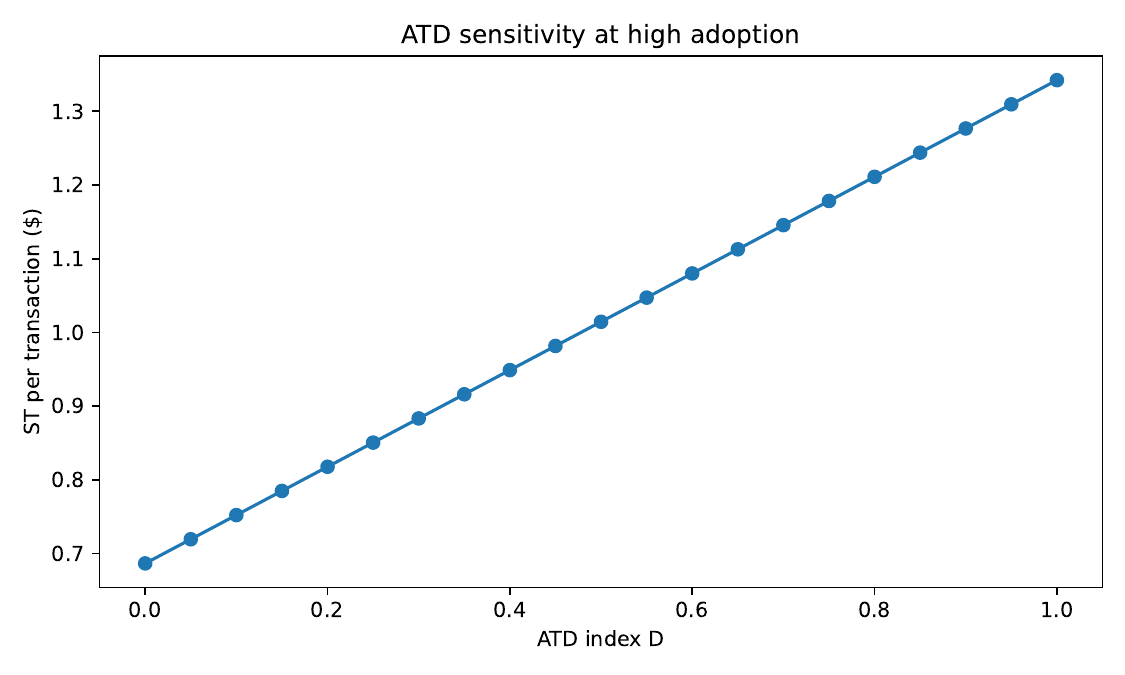}
\caption{ATD sensitivity sweep at a fixed high-adoption operating point.}
\label{fig:atd_sensitivity}
\end{figure}

\subsection{Twelve-Month Dynamic Paths}

The third simulation shows how governance changes the path. Both workflows start at the same debt level and the same high-adoption operating point. Path A accumulates more debt than it remediates. Path B remediates more debt than it accumulates.

For this illustration, we use the simplified debt dynamics in Equation~\ref{eq:debt_dynamics_simple}. Path A uses $\Delta^{acc}-\Delta^{rem}=0.06-0.02=+0.04$ per month. Path B uses $\Delta^{acc}-\Delta^{rem}=0.03-0.07=-0.04$ per month. The companion spreadsheet exposes all four rates as editable inputs.

\begin{table}[H]
\centering
\caption{Twelve-Month Evolution of \ATD{} and \ST{} per Transaction}
\label{tab:dynamic_paths}
\small
\begin{tabularx}{\textwidth}{rrrrr}
\toprule
\textbf{Month} & \textbf{Path A: $D$} & \textbf{Path A: $\overline{ST}$} & \textbf{Path B: $D$} & \textbf{Path B: $\overline{ST}$} \\
\midrule
0 & 0.10 & \$0.75 & 0.10 & \$0.75 \\
3 & 0.22 & \$0.83 & 0.00 & \$0.69 \\
6 & 0.34 & \$0.91 & 0.00 & \$0.69 \\
9 & 0.46 & \$0.99 & 0.00 & \$0.69 \\
12 & 0.58 & \$1.07 & 0.00 & \$0.69 \\
\bottomrule
\end{tabularx}
\end{table}

By month 12, the accumulation path pays about \$0.38 more per transaction than the governance path. At 10,000 transactions per month, that is about \$3,804 per month, or roughly \$45,649 per year, for one workflow. The point is not that the illustrative dollar values are universal. The point is that the model turns a vague concern about debt into a dashboard quantity that can be tracked, simulated, and governed.

\begin{figure}[H]
\centering
\includegraphics[width=0.82\textwidth]{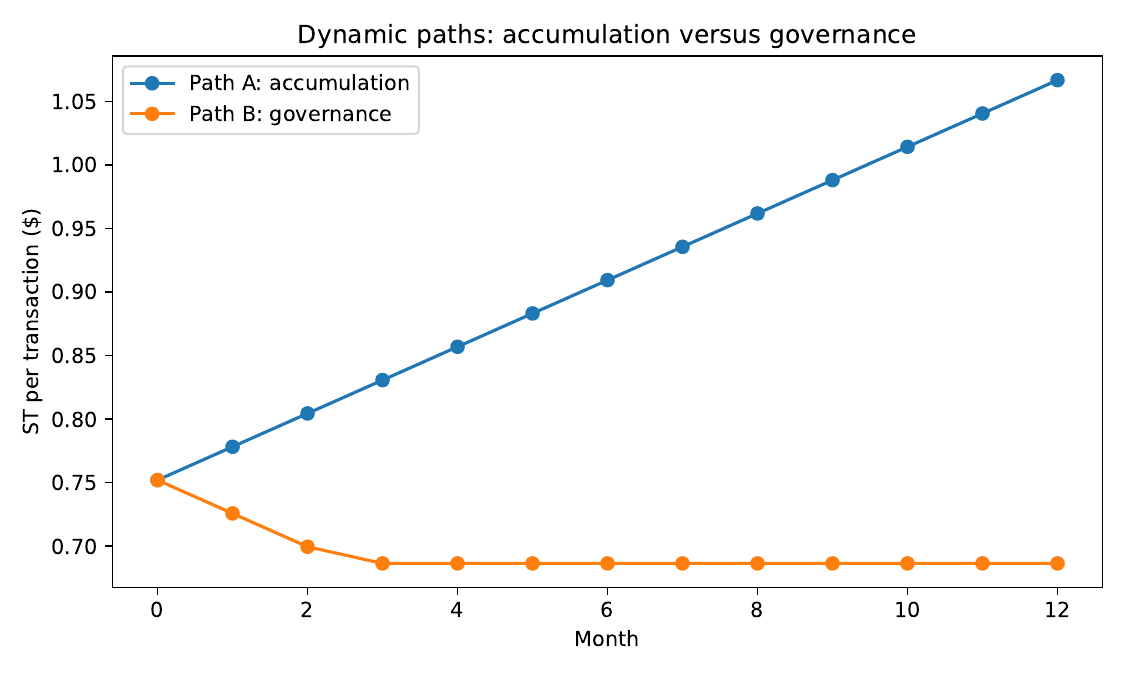}
\caption{Dynamic paths for an accumulation regime and a governance regime.}
\label{fig:dynamic_paths}
\end{figure}

\section{From Model to Dashboard}

The model supports a managerial dashboard that answers four recurring questions.

\begin{enumerate}
    \item \textbf{Is operating burden stable, rising, or concentrated?} Track $TST_{w,t}$ and $\overline{ST}_{w,t}$ by workflow and period.
    \item \textbf{Is rising tax explained by adoption, surface area, autonomy, variability, or debt?} Decompose changes in tax across drivers rather than treating every increase as debt.
    \item \textbf{Is debt accumulating or being paid down?} Track $D_{w,t}$ and its six components over time.
    \item \textbf{Which intervention is appropriate?} Use the decomposition to distinguish refactoring decisions from operating-control decisions.
\end{enumerate}

Table~\ref{tab:dashboard_fields} gives a minimal dashboard design.

\begin{table}[H]
\centering
\caption{Dashboard Fields per Workflow per Period}
\label{tab:dashboard_fields}
\small
\begin{tabularx}{\textwidth}{p{0.20\textwidth}p{0.35\textwidth}X}
\toprule
\textbf{Field} & \textbf{Definition} & \textbf{Managerial use} \\
\midrule
$D_{w,t}$ & Composite \ATD{} index in $[0,1]$, often reported as 0 to 100 & Track debt stock and trigger governance review above threshold. \\
Component scores $d^i_{w,t}$ & Debt scores for context, tools, memory, orchestration, observability, and platform coupling & Identify the remediation target. \\
$TST_{w,t}$ & Total stochastic tax for the period & Budget item and operating-cost allocation. \\
$\overline{ST}_{w,t}$ & Tax per completed transaction & Compare unit economics across workflows. \\
$\overline{ST}^{0}_{w,t}$ & Baseline tax when $D=0$ at current operating exposure & Set expectations for debt-paydown benefits, subject to calibration uncertainty. \\
$\overline{ST}^{D}_{w,t}$ & Debt-amplified tax & Estimate avoidable burden associated with accumulated debt, after $\beta_k$ values are calibrated. \\
Driver indicators & Adoption, surface area, horizon, autonomy, variability & Diagnose whether tax growth is due to scale, complexity, or debt. \\
Calibration status & Expert prior, before-after estimate, regression estimate, or validated estimate & Prevents the dashboard from overstating precision. \\
\bottomrule
\end{tabularx}
\end{table}

The dashboard should also display category diagnostics such as token cost per transaction, retry rate, escalation rate, P95 latency, security-flag rate, and revalidation hours after changes. These signals help managers move from measurement to action. If token cost rises while debt is flat, context-management or retrieval strategy may be the right target. If retry and revalidation costs rise together with tool/schema debt, refactoring interfaces may be more important. If total tax rises while per-transaction tax falls, the system may be scaling responsibly.

\section{Implementation Guidance}

A practical implementation can proceed in seven steps.

\begin{enumerate}
    \item \textbf{Define workflow boundaries.} Specify where the workflow begins and ends, such as invoice intake to payment scheduling.
    \item \textbf{Score the debt stock.} Score the six debt components using evidence from prompts, tools, traces, tests, release records, and governance artifacts.
    \item \textbf{Collect operating signals.} Capture token use, retry rate, escalation rate, latency, security flags, evaluation effort, monitoring effort, and revalidation work.
    \item \textbf{Convert signals into dollars.} Use token prices, labor rates, delay costs, tool-call charges, and allocated fixed costs.
    \item \textbf{Calibrate the structural parameters.} Estimate $F_k$ and $V_k$ from accounting and logs. Estimate $\gamma$ values from scale, surface, horizon, autonomy, and variability changes. Treat $\beta_k$ as an expert prior until the team has before-after or cross-workflow evidence.
    \item \textbf{Estimate the baseline with uncertainty.} Calculate $\overline{ST}^{0}_{w,t}$ by setting $D_{w,t}=0$ only after choosing a calibrated or sensitivity-tested set of $\beta_k$ values.
    \item \textbf{Use the decomposition for decisions.} Treat baseline tax as an operating reality to manage and debt-amplified tax as a candidate for remediation.
\end{enumerate}

The model is feasible because it does not require perfect causal identification before it becomes useful. A dashboard can begin with observed costs and expert-calibrated parameters, then improve as the organization observes releases, incidents, growth, and remediation efforts.

\section{Limitations and Extensions}

This framework is intended for conceptual clarification and managerial measurement. It is not a universal metric. Several limitations should be acknowledged.

First, cost attribution can be imperfect. Evaluation, monitoring, and security costs are often shared across workflows. Managers should use transparent allocation rules and avoid false precision. Second, the causal effect of \ATD{} on \ST{} may be difficult to identify without interventions, redesigns, or controlled comparisons. The $\beta_k$ coefficients should therefore be treated as calibrated estimates, not directly observed truths. Third, rare but severe failures may not be well represented by average costs. A dashboard should include tail-risk indicators in addition to average tax. Fourth, some organizations may intentionally accept a high baseline tax in regulated or safety-critical settings because stronger evaluation, monitoring, and human review are prudent. Fifth, functional forms may need adjustment. Some systems may exhibit thresholds, nonlinear cascades, or saturation effects that the linear debt amplifier does not capture. Sixth, the model in this note treats $\overline{ST}_{w,t}$ as a deterministic function of its drivers, even though underlying event rates such as $q^j_{w,t}$ are random variables. A natural extension is to model $\overline{ST}_{w,t}$ as an expected value with a variance derived from event-rate distributions, which would let the second meaning of \ST{} be displayed as a confidence band on the dashboard rather than a point estimate.

These limitations do not weaken the usefulness of the model. They clarify how it should be used. The model is a disciplined accounting and governance lens, not a replacement for managerial judgment.

\section{Conclusion}

\ATD{} and \ST{} are best modeled as distinct but connected properties of agentic systems. \ATD{} is the stock of accumulated design and governance liability. \ST{} is the flow of recurring operating burden created by stochastic action in real workflows. Debt can amplify the tax, but the tax also arises from adoption, surface area, workflow horizon, autonomy, and inherent model variability. The model in this note makes that distinction operational. It gives managers a debt index, a cost-category structure, a baseline versus debt-amplified decomposition, a calibration path for the difficult $\beta_k$ parameters, and a simulation approach that can be implemented in a spreadsheet dashboard. This is why tax is a better metaphor than interest: interest suggests a rate paid on a fixed debt principal, whereas \ST{} includes a persistent baseline and a rate that can move with debt, usage, surface area, autonomy, and the stochastic event process itself.

\appendix

\section{Full Simulation Parameters}
\label{app:params}

Reference values used in the simulation are $U_0=1{,}000$, $S_0=5$, $H_0=4$, $A_0=0.40$, and $\Theta_0=0.30$. Table~\ref{tab:full_params} reports all amplifier coefficients, including $\gamma^\Theta_k$ as a column.

\begin{longtable}{p{0.22\textwidth}rrrrrrrr}
\caption{Full Cost-Category Parameters}\label{tab:full_params}\\
\toprule
\textbf{Category} & $V_k$ & $F_k$ & $\beta_k$ & $\gamma^U_k$ & $\gamma^S_k$ & $\gamma^H_k$ & $\gamma^A_k$ & $\gamma^\Theta_k$ \\
\midrule
\endfirsthead
\toprule
\textbf{Category} & $V_k$ & $F_k$ & $\beta_k$ & $\gamma^U_k$ & $\gamma^S_k$ & $\gamma^H_k$ & $\gamma^A_k$ & $\gamma^\Theta_k$ \\
\midrule
\endhead
Evaluation & 0.010 & 200 & 1.2 & 0.12 & 0.08 & 0.08 & 0.12 & 0.15 \\
Monitoring & 0.005 & 350 & 0.4 & 0.10 & 0.06 & 0.04 & 0.06 & 0.08 \\
Retry and repair & 0.040 & 0 & 2.5 & 0.07 & 0.05 & 0.12 & 0.10 & 0.12 \\
Escalation & 0.120 & 80 & 1.0 & 0.08 & 0.06 & 0.10 & 0.18 & 0.08 \\
Revalidation & 0.002 & 400 & 1.8 & 0.10 & 0.12 & 0.12 & 0.10 & 0.20 \\
Latency and delay & 0.030 & 50 & 0.8 & 0.05 & 0.05 & 0.18 & 0.15 & 0.10 \\
Token, compute, and context & 0.200 & 0 & 0.6 & 0.03 & 0.08 & 0.12 & 0.08 & 0.08 \\
Security and guardrails & 0.008 & 150 & 0.3 & 0.15 & 0.18 & 0.05 & 0.20 & 0.12 \\
\bottomrule
\end{longtable}

\section{Equations in One Place}
\label{app:equations}

For reference, the central model consists of the following equations:

\begin{align}
\mathbf{d}_{w,t} &= \left(d^{ctx}_{w,t}, d^{tool}_{w,t}, d^{mem}_{w,t}, d^{orch}_{w,t}, d^{obs}_{w,t}, d^{plat}_{w,t}\right), \tag{Debt components} \\
 d^i_{w,t+1} &= \clip\left[(1-\phi_i R^i_{w,t})d^i_{w,t}+\alpha_i X^i_{w,t}Q^i_{w,t}(1-G^i_{w,t})+\zeta_i V^{plat}_{w,t}E^i_{w,t}\right], \tag{Debt dynamics} \\
D_{w,t} &= \clip\left[\sum_i\omega_i d^i_{w,t}+\sum_{i<j}\omega_{ij}d^i_{w,t}d^j_{w,t}\right], \tag{Debt index} \\
C^k_{w,t} &= \left(F_k+V_kN_{w,t}\right)\Phi_k(D_{w,t})\Psi_k(U_{w,t},S_{w,t},H_{w,t},A_{w,t},\Theta_{w,t}), \tag{Cost category} \\
\Phi_k(D_{w,t}) &= 1+\beta_kD_{w,t}, \tag{Debt amplifier} \\
\Psi_k(\cdot) &= \exp\Bigg[\gamma^U_k\ln\left(\frac{U_{w,t}}{U_0}\right)+\gamma^S_k\ln\left(\frac{S_{w,t}}{S_0}\right)+\gamma^H_k\ln\left(\frac{H_{w,t}}{H_0}\right) \nonumber \\
&\quad +\gamma^A_k(A_{w,t}-A_0)+\gamma^\Theta_k(\Theta_{w,t}-\Theta_0)\Bigg], \tag{Exposure amplifier} \\
TST_{w,t} &= \sum_{k\in\mathcal{K}} C^k_{w,t}, \qquad \overline{ST}_{w,t}=\frac{TST_{w,t}}{N_{w,t}}, \tag{Total and average ST} \\
\overline{ST}_{w,t} &= \overline{ST}^{0}_{w,t}+\overline{ST}^{D}_{w,t}. \tag{Baseline plus debt-amplified ST}
\end{align}

\section{Companion Excel Spreadsheet}
\label{app:workbook}

The companion Excel spreadsheet implements the same notation and model as this note. It is designed to let readers change reference values, category parameters, scenario inputs, and dynamic-path assumptions while seeing how the stochastic-tax dashboard responds. The spreadsheet contains:

\begin{itemize}
    \item \textbf{README:} a concise explanation of the spreadsheet's purpose and notation.
    \item \textbf{Parameters:} editable reference values, cost-category parameters, six ATD components, scenario inputs, and dynamic-path inputs.
    \item \textbf{Scenarios:} formula-driven category-cost calculations for the four scenarios.
    \item \textbf{Summary:} total and per-transaction tax, baseline tax, debt-amplified tax, and key computed insights.
    \item \textbf{ATD\_Sensitivity:} a sweep from $D=0$ to $D=1$ at a fixed high-adoption operating point.
    \item \textbf{Time\_Series:} the accumulation and governance paths using the simplified debt dynamics in Equation~\ref{eq:debt_dynamics_simple}.
    \item \textbf{Data\_Dictionary:} definitions of all symbols used in the spreadsheet.
    \item \textbf{Dashboard:} a compact managerial view of scenario results and key performance indicators.
\end{itemize}

The spreadsheet is intended for exploration, not as a universal benchmark. Managers should replace the illustrative parameters with their own token prices, labor rates, event rates, observed category costs, debt scores, and calibration assumptions.

\end{document}